\documentclass[conference]{IEEEtran}
\IEEEoverridecommandlockouts
\usepackage{multicol}
\usepackage{multirow} 
\usepackage{cite}
\usepackage{amsmath,amssymb,amsfonts}
\usepackage{algorithmic}
\usepackage{graphicx}
\usepackage{float}
\usepackage{textcomp}
\usepackage{hyperref}
\usepackage{xcolor}
\usepackage{booktabs}
\usepackage{bm}
\usepackage[table]{xcolor}
\def\BibTeX{{\rm B\kern-.05em{\sc i\kern-.025em b}\kern-.08em
    T\kern-.1667em\lower.7ex\hbox{E}\kern-.125emX}}
\begin{document}

\title{Multi-Perspective Subimage CLIP with Keyword Guidance for Remote Sensing Image-Text Retrieval}

\author{
    \IEEEauthorblockN{
        Yifan Li\textsuperscript{1}, 
        Shiying Wang\textsuperscript{1,2,$\dagger$}, 
        Jianqiang Huang\textsuperscript{1,2,$\dagger$}
    }
    \IEEEauthorblockA{
        \textit{\textsuperscript{1}School of Computer Technology and Applications, Qinghai University, Xining, China}\\
        \textit{\textsuperscript{2}Qinghai Provincial Laboratory for Intelligent Computing and Application, Xining, China}
    }
}

\maketitle


\begin{abstract}
Vision-Language Pre-training (VLP) models like CLIP have significantly advanced Remote Sensing Image-Text Retrieval (RSITR). However, existing methods predominantly rely on coarse-grained global alignment, which often overlooks the dense, multi-scale semantics inherent in overhead imagery. Moreover, adapting these heavy models via full fine-tuning incurs prohibitive computational costs and risks catastrophic forgetting. To address these challenges, we propose \textit{MPS-CLIP}, a parameter-efficient framework designed to shift the retrieval paradigm from global matching to keyword-guided fine-grained alignment. Specifically, we leverage a Large Language Model (LLM) to extract core semantic keywords, guiding the Segment Anything Model (SamGeo) to generate semantically relevant sub-perspectives. To efficiently adapt the frozen backbone, we introduce a Gated Global Attention (G$^2$A) adapter, which captures global context and long-range dependencies with minimal overhead. Furthermore, a Multi-Perspective Representation (MPR) module aggregates these local cues into robust multi-perspective embeddings. The framework is optimized via a hybrid objective combining multi-perspective contrastive and weighted triplet losses, which dynamically selects maximum-response perspectives to suppress noise and enforce precise semantic matching. Extensive experiments on the RSICD and RSITMD benchmarks demonstrate that MPS-CLIP achieves state-of-the-art performance with 35.18\% and 48.40\% mean Recall (mR), respectively, significantly outperforming full fine-tuning baselines and recent competitive methods. Code is available at \url{https://github.com/Lcrucial1f/MPS-CLIP}.
\end{abstract}


\section{Introduction}
\label{sec:intro}

Remote Sensing Image-Text Retrieval (RSITR) bridges vision and language modalities for retrieval tasks, underpinning applications such as urban planning\cite{ma2015remote}. However, distinct from natural scenes, remote sensing imagery features overhead perspectives and dense object distributions. Coupled with spatially complex textual descriptions, these attributes hinder existing methods from achieving fine-grained semantic alignment.

Early RSITR approaches primarily relied on training-from-scratch dual-stream architectures, adopting CNN backbones \cite{yuan2021lightweight,chen2023multiscale,pan2023reducing,wu2024spatial} and Transformer-style designs \cite{yuan2022exploring,tang2023interacting,ma2024direction,yang2024transcending,he2024visual,ji2023knowledge} to extract visual and textual features. These representations were then aligned in a shared latent space via metric-learning objectives \cite{yuan2021lightweight,chen2023multiscale,yuan2022exploring}. However, such methods were severely constrained by the limited scale of labeled RS datasets and the prevalent from-scratch training paradigm \cite{yuan2021lightweight,chen2023multiscale,ma2024direction}. As a result, models were prone to overfitting and often struggled to capture intricate cross-modal semantic interactions in complex remote-sensing scenes \cite{tang2023interacting,wu2024spatial,he2024visual}.

Recently, large-scale Vision-Language Pre-training models, represented by CLIP \cite{radford2021learning}, have revolutionized the field with their exceptional zero-shot transfer capabilities. Despite this progress, most existing CLIP-based RS methods still largely inherit the standard global image-text alignment paradigm \cite{radford2021learning,wang2024skyscript,lu2023uniadapter,yuan2023parameter,huang2024efficient,zou2025adapting}. While effective for natural images, this coarse-grained approach is ill-suited for RS imagery, where key semantics are often embedded in dense objects and their spatial relationships \cite{chen2023multiscale,pan2023reducing,wu2024spatial,zou2024diffcr}. By relying solely on global feature embeddings, these methods frequently overlook fine-grained local cues, making it difficult to distinguish hard samples that share similar backgrounds but differ in specific entity layouts \cite{pan2023reducing,wu2024spatial}. Moreover, adapting these heavy models creates a secondary challenge: the common full-parameter fine-tuning strategy incurs high computational costs and poses significant risks of overfitting and catastrophic forgetting, which motivates parameter-efficient adaptation strategies \cite{gao2024clip,huang2024efficient,zou2025dynamic,li2024subnetwork,li2025apollo}.

\begin{figure*}[t]
    \centering
    \includegraphics[width=0.90\linewidth]{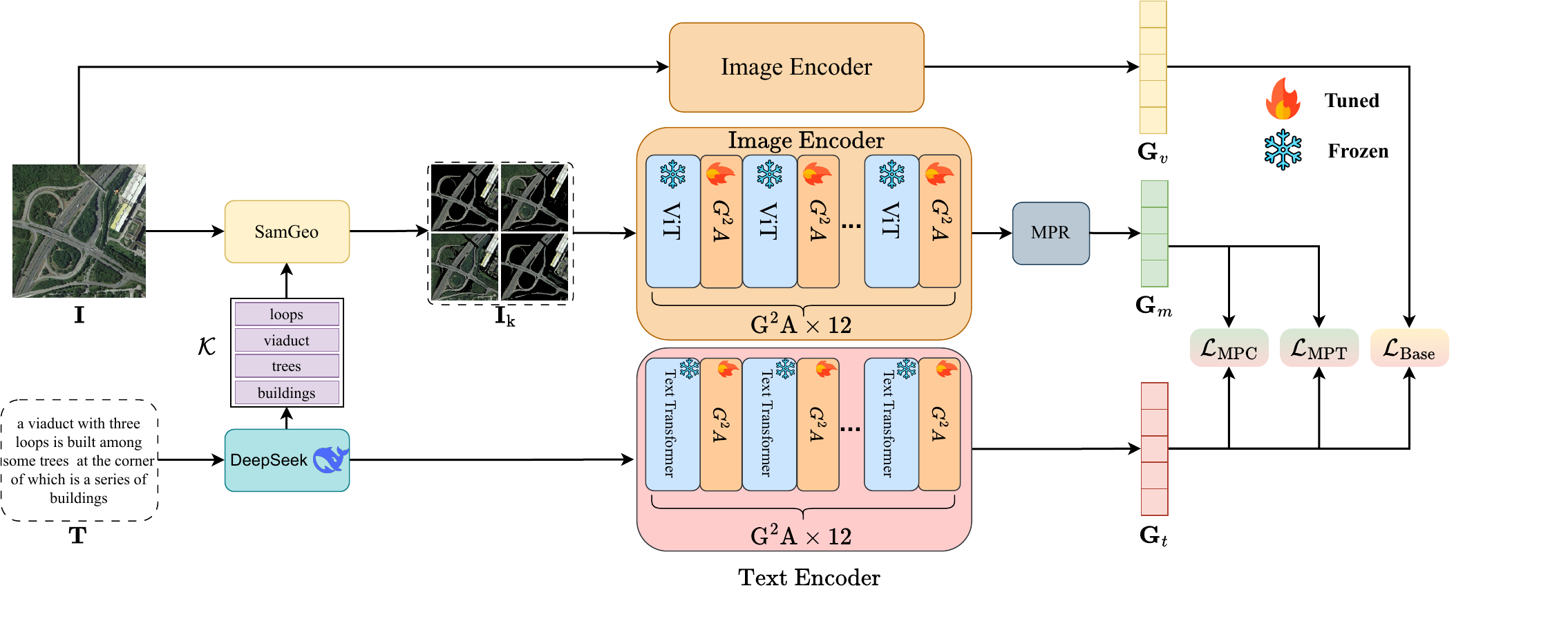}
    \caption{The overall framework of MPS-CLIP. The architecture integrates DeepSeek and SamGeo for semantic-aware patch generation.}
    \label{fig:framework}
    \vspace{-15pt}
\end{figure*}

To address these challenges, we propose MPS-CLIP, a keyword-guided multi-perspective framework designed to shift the retrieval paradigm from coarse-grained global matching to semantic-driven fine-grained alignment, as shown in Fig.  \ref{fig:framework}. Specifically, we leverage a LLM to mine core semantic keywords, such as object categories and spatial relationships, which guide SamGeo \cite{wu2023samgeo} to generate semantically relevant sub-perspectives. For feature extraction, we employ CLIP with a novel Gated Global Attention  (G$^2$A) adapter. This adapter utilizes the [CLS] token to efficiently capture global semantics with minimal computational overhead. Subsequently, a Multi-Perspective Representation (MPR) module performs deep fusion of global and local features. Finally, the framework is trained via a hybrid objective combining multi-perspective contrastive loss and weighted triplet loss to ensure robust feature alignment.

We conducted extensive experiments on the RSICD and RSITMD benchmarks. Results demonstrate that MPS-CLIP achieves state-of-the-art (SOTA) performance with 35.18\% and 48.40\% mR, respectively, significantly outperforming recent competitive methods and full fine-tuning baselines.

\section{Related Work}

\subsection{Training-from-Scratch Methods}

Early RSITR approaches predominantly relied on dual-stream architectures training from scratch, employing CNNs and Transformer architectures to map visual and textual features into a shared space via metric learning. To address the unique challenges of remote sensing, such as scale variations and complex backgrounds, extensive research has focused on optimizing feature interaction and attention mechanisms. For instance, methods like LW-MCR  \cite{yuan2021lightweight}, AMFMN \cite{yuan2022exploring}, and MSA \cite{yang2024transcending}introduced multiscale self-attention and dynamic filtering to enhance feature robustness. Moving towards fine-grained alignment, MSITA \cite{chen2023multiscale}, SWAN \cite{pan2023reducing}, and SCAT-PRG \cite{wu2024spatial} utilized techniques ranging from image-guided text alignment to pseudo-region generation to capture local details. Furthermore, to mitigate semantic ambiguity, works such as DOVE \cite{ma2024direction}, IEFT \cite{tang2023interacting}, VGSGN \cite{he2024visual}, and KAMCL \cite{ji2023knowledge} incorporated prior knowledge and deep cross-modal interactions. Despite these architectural innovations, these methods are fundamentally constrained by their reliance on limited annotated datasets, often resulting in restricted generalization capabilities compared to modern pre-training paradigms.

\subsection{Pre-training and Adaptation Methods}
The emergence of CLIP \cite{radford2021learning} has revolutionized retrieval tasks with its powerful zero-shot transferability. To bridge the domain gap between natural and remote sensing scenes, methods like SkyCLIP \cite{wang2024skyscript} and SingleLanguage \cite{al2022multilanguage} explored continual pre-training and multilingual adaptation. However, given the prohibitive costs of full fine-tuning, Parameter-Efficient Fine-Tuning (PEFT) has become the mainstream solution. General approaches like CLIP-Adapter \cite{gao2024clip} and Cross-modal Adapter \cite{jiang2022cross} insert lightweight bottleneck layers, while RS-specific methods such as UniAdapter \cite{lu2023uniadapter}, PE-RSITR  \cite{yuan2023parameter}, and HarMA  \cite{huang2024efficient} further optimize efficiency through weight sharing or hierarchical designs. Although these adapter-based methods effectively reduce computational overhead, they largely inherit the coarse-grained global alignment paradigm of CLIP. Consequently, they often overlook the complex spatial structures and fine-grained local semantics inherent in RS imagery, a limitation our proposed keyword-guided multi-perspective framework explicitly addresses.

\section{Methods}

\subsection{Overall Framework}

As illustrated in Fig.  \ref{fig:framework}, MPS-CLIP builds upon the global alignment paradigm of CLIP by introducing a multi-perspective modeling mechanism that links text keywords to image sub-perspectives. The framework primarily consists of three core components: the Gated Global Attention Adapter (G$^2$A), the Multi-Perspective Representation (MPR) module, and the multi-perspective loss function. Through this design, MPS-CLIP effectively enhances fine-grained semantic alignment capabilities in remote sensing scenarios.

Specifically, given input text $\mathbf{T} \in \mathbb{R}^L$ (where $L$ denotes sequence length) and input image $\mathbf{I} \in \mathbb{R}^{C \times H \times W}$ (where $C$ is the number of channels, and $H$ and $W$ are the height and width), MPS-CLIP adopts distinct feature extraction strategies for text and images. On the text side, we utilize the CLIP text encoder fine-tuned with the G$^2$A adapter to extract global text features $\mathbf{G}_{t} \in \mathbb{R}^D$, where $D$ indicates the feature dimension.

On the image side, the feature extraction process is divided into global and local branches. In the global branch, we employ the CLIP visual encoder fine-tuned with the G$^2$A adapter to obtain global visual features $\mathbf{G}_{v} \in \mathbb{R}^D$. In the local branch, we first use a LLM (\texttt{Deepseek V3.2} \cite{liu2024deepseek}) to extract keywords $\mathcal{K}$ from text $\mathbf{T}$ and guide SamGeo \cite{wu2023samgeo} to generate a set of semantically relevant sub-perspectives $\{\mathbf{I}_k \mid k = 1, 2, \dots, K\}$. Subsequently, these sub-perspectives are processed by the aforementioned fine-tuned CLIP visual encoder to yield local visual features $\mathbf{G}_{l} \in \mathbb{R}^{D \times K}$, where $K$ represents the number of sub-perspectives. Next, we input the local features $\mathbf{G}_{l}$ into the MPR module to generate multi-perspective visual features $\mathbf{G}_{m} \in \mathbb{R}^{D \times K}$.

Finally, the model outputs the globally aligned feature pair $(\mathbf{G}_{v}, \mathbf{G}_{t})$ and the fine-grained interactive feature set $\mathbf{G}_{m}$ to achieve multi-level cross-modal matching.

\subsection{CLIP Encoder with the G$^2$A Adapter}
\label{sec:clip_g2a}

To improve the transferability of global representations and multi-perspective interaction capability in remote sensing scenarios, we insert the G$^2$A adapter into the Transformer layers of CLIP. We freeze the backbone parameters and update only the G$^2$A adapter parameters to mitigate catastrophic forgetting. Let the input token features of a Transformer layer be $\mathbf{x}\in\mathbb{R}^{N\times D}$, where $N$ denotes the number of tokens. In the vision branch, $N$ corresponds to the number of spatial patch tokens and an optional [CLS] token, while in the text branch it corresponds to the text sequence length.

\begin{figure}[H]
    \centering
    \includegraphics[width=0.85\linewidth]{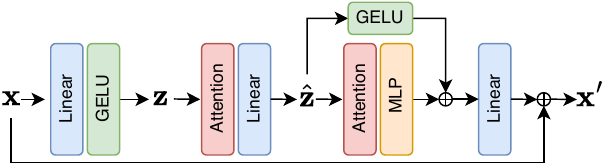}
    \caption{Architecture of the proposed G$^2$A adapter.}
    \label{fig:g2a_adapter}
\end{figure}

As shown in Fig.  \ref{fig:g2a_adapter}, given the input token features $\mathbf{x}\in\mathbb{R}^{N\times D}$,
the proposed G$^2$A adapter first projects $\mathbf{x}$ into a compact subspace via a bottleneck
transformation, followed by global interaction in the low-dimensional space.
Specifically, we obtain the compressed representation by
\begin{equation}
\mathbf{z}=\phi(\mathbf{x}\mathbf{W}_{1}+\mathbf{b}_{1}), \qquad \mathbf{z}\in\mathbb{R}^{N\times d},
\end{equation}
where $\mathbf{W}_{1}\in\mathbb{R}^{D\times d}$, $\mathbf{b}_{1}\in\mathbb{R}^{d}$ are learnable parameters,
$d\ll D$, and $\phi(\cdot)$ denotes the GELU activation.
Next, we perform multi-head self-attention on $\mathbf{z}$ and apply a linear transformation:
\begin{equation}
\mathbf{z}_{\mathrm{attn}}=\mathrm{Attn}(\mathbf{z}), \qquad \mathbf{z}_{\mathrm{attn}}\in\mathbb{R}^{N\times d},
\end{equation}
\begin{equation}
\hat{\mathbf{z}}=\mathbf{z}_{\mathrm{attn}}\mathbf{W}_{2}+\mathbf{b}_{2}, \qquad \hat{\mathbf{z}}\in\mathbb{R}^{N\times d},
\end{equation}
where $\mathrm{Attn}(\cdot)$ is the standard multi-head self-attention operator over $d$ channels,
$\mathbf{W}_{2}\in\mathbb{R}^{d\times d}$ and $\mathbf{b}_{2}\in\mathbb{R}^{d}$.
On top of $\hat{\mathbf{z}}$, the adapter further enhances long-range dependency modeling by
combining a global attention branch and a feed-forward branch, and keeps the output dimension unchanged
to support residual learning:
\begin{equation}
\widetilde{\mathbf{z}}=\hat{\mathbf{z}}+\mathrm{MLP}(\mathrm{Attn}(\hat{\mathbf{z}})),
\qquad \widetilde{\mathbf{z}}\in\mathbb{R}^{N\times d},
\end{equation}
where $\mathrm{MLP}(\cdot)$ denotes a position-wise feed-forward network mapping
$\mathbb{R}^{N\times d}\rightarrow\mathbb{R}^{N\times d}$.
To adaptively control the contribution of the enhanced features, we introduce a learnable scalar gate:
\begin{equation}
\mathbf{z}_{\mathrm{gate}}=\sigma(\gamma)\,\widetilde{\mathbf{z}}, \qquad \mathbf{z}_{\mathrm{gate}}\in\mathbb{R}^{N\times d},
\end{equation}
where $\gamma$ is a learnable parameter and $\sigma(\cdot)$ is the Sigmoid function. Finally, the gated features are projected back to the original feature dimension and added to the input via
a residual connection:
\begin{equation}
\mathbf{x}_{\mathrm{up}}=\mathbf{z}_{\mathrm{gate}}\mathbf{W}_{3}+\mathbf{b}_{3}, \qquad \mathbf{x}_{\mathrm{up}}\in\mathbb{R}^{N\times D},
\end{equation}
\begin{equation}
\mathbf{x}'=\mathbf{x}+\mathbf{x}_{\mathrm{up}}, \qquad \mathbf{x}'\in\mathbb{R}^{N\times D},
\end{equation}
where $\mathbf{W}_{3}\in\mathbb{R}^{d\times D}$ and $\mathbf{b}_{3}\in\mathbb{R}^{D}$ are learnable parameters.

\subsection{Multi-Perspective Representation (MPR) Module}

In remote sensing, keyword-guided sub-perspectives $\{\mathbf{I}_k\}_{k=1}^{K}$ contain both complementary cues and redundant noise, so independently aligning each sub-perspective is sensitive to scale changes, background clutter, and segmentation bias, hurting fine-grained semantic matching. We therefore propose the Multi-Perspective Representation (MPR) module, which aggregates correlated sub-perspectives into a robust local semantic summary, then projects it into complementary multi-perspective embeddings to support keyword-driven multi-granularity interaction alignment. MPR outputs multi-perspective visual features $\mathbf{G}_{m}$ for fine-grained matching.

Specifically, we first aggregate $\mathbf{G}_{l}$ to obtain a local semantic summary $\mathbf{e}\in\mathbb{R}^{D}$:
\begin{equation}
\mathbf{e}=\frac{1}{K}\sum_{k=1}^{K}\mathbf{G}_{l,k},
\end{equation}
where $\mathbf{G}_{l,k}$ denotes the $k$th sub perspective vector of $\mathbf{G}_{l}$. Next, the MPR module maps $\mathbf{e}$ into $K$ complementary semantic subspaces using $K$ two layer MLP heads with independent parameters (Linear, GELU, Linear, optional Dropout). The $k$th head is computed as
\begin{equation}
\mathbf{v}_k' = \mathbf{W}_{k,2}\,\phi(\mathbf{W}_{k,1}\mathbf{e} + \mathbf{b}_{k,1}) + \mathbf{b}_{k,2},
\quad k=1,\dots,K,
\end{equation}
where $\mathbf{W}_{k,1}$ and $\mathbf{W}_{k,2}$ are the weight matrices, and $\mathbf{b}_{k,1}$ and $\mathbf{b}_{k,2}$ are the bias vectors of the $k$th MLP head. We then apply $L_2$ normalization to each perspective vector to improve comparability and numerical stability in contrastive learning:
\begin{equation}
\mathbf{v}_k = \frac{\mathbf{v}_k'}{\|\mathbf{v}_k'\|_2 + \varepsilon},
\quad k=1,\dots,K.
\end{equation}
Finally, the output multi-perspective visual representation is
\begin{equation}
\mathbf{G}_{m}=\{\mathbf{v}_1,\dots,\mathbf{v}_K\}.
\end{equation}

\begin{table*}[t]
\centering
\caption{Comparisons of retrieval performance on RSICD and RSITMD test sets.
\textbf{Bold} indicates the best result and \underline{underlined} indicates the second best.}
\label{tab:rsicd_rsitmd}
\setlength{\tabcolsep}{3.2pt}

\begin{tabular}{l|ccc|ccc|c|ccc|ccc|c}
\toprule
\multirow{3}{*}{\textbf{Methods}} &
\multicolumn{7}{c|}{\textbf{RSICD}} &
\multicolumn{7}{c}{\textbf{RSITMD}} \\
\cline{2-15} 
& \multicolumn{3}{c|}{\textbf{Text Retrieval}} &
  \multicolumn{3}{c|}{\textbf{Image Retrieval}} &
  \multirow{2}{*}{\textbf{mR}} &
  \multicolumn{3}{c|}{\textbf{Text Retrieval}} &
  \multicolumn{3}{c|}{\textbf{Image Retrieval}} &
  \multirow{2}{*}{\textbf{mR}} \\
\cline{2-7}\cline{9-14} 
& \textbf{R@1} & \textbf{R@5} & \textbf{R@10} &
  \textbf{R@1} & \textbf{R@5} & \textbf{R@10} & &
  \textbf{R@1} & \textbf{R@5} & \textbf{R@10} &
  \textbf{R@1} & \textbf{R@5} & \textbf{R@10} & \\
\midrule

\multicolumn{15}{c}{\textit{\textbf{Train from Scratch}}} \\
\midrule

KAMCL\cite{ji2023knowledge}       & 11.99 & 27.17 & 38.33 & 7.78 & 25.23 & 40.02 & 25.08 &
                    16.81 & 34.96 & 47.12 & 14.60 & 41.86 & 59.86 & 35.87 \\
VGSGN\cite{he2024visual}       & 8.33  & 21.87 & 32.57 & 6.53 & 23.13 & 36.85 & 21.55 &
                    14.16 & 34.96 & 50.66 & 13.23 & 42.57 & 63.41 & 36.50 \\
DOVE\cite{ma2024direction}        & 8.66  & 22.35 & 34.95 & 6.04 & 23.95 & 40.35 & 22.72 &
                    16.81 & 36.80 & 50.93 & 12.20 & 49.93 & 66.50 & 37.73 \\
MSA\cite{yang2024transcending}         & 9.52  & 25.25 & 35.32 & 6.55 & 24.43 & 39.63 & 23.45 &
                    15.93 & 38.50 & 50.88 & 14.96 & 45.22 & 61.86 & 37.89 \\
PIR-ITR\cite{pan2024pir}     & 10.89 & 26.17 & 37.79 & 7.17 & 25.07 & 41.06 & 24.69 &
                    18.36 & 42.04 & 55.53 & 13.36 & 44.47 & 61.73 & 39.25 \\
\midrule

\multicolumn{15}{c}{\textit{\textbf{CLIP-based}}} \\
\midrule

SkyCLIP\cite{wang2024skyscript}       & 6.59 & 16.10 & 26.53 & 7.14 & 22.34 & 34.29 & 18.83 &
                      10.18 & 25.44 & 35.62 & 10.88 & 33.27 & 49.82 & 27.54 \\
CLIP-Adapter\cite{gao2024clip}    & 7.11 & 19.48 & 31.01 & 7.67 & 24.87 & 39.73 & 21.65 &
                      12.83 & 28.84 & 39.05 & 13.30 & 40.20 & 60.06 & 32.38 \\
Linear-probe CLIP\cite{radford2021learning}   & 8.46 & 24.41 & 37.72 & 7.81 & 25.89 & 42.47 & 24.46 &
                      17.02 & 33.12 & 48.35 & 13.33 & 41.80 & 63.89 & 36.25 \\
UniAdapter\cite{lu2023uniadapter}      & 12.65 & 30.81 & 42.74 & 9.61 & 30.06 & 47.16 & 28.84 &
                      19.86 & 36.32 & 51.28 & 17.54 & 44.89 & 56.46 & 39.23 \\
SingleLanguage\cite{al2022multilanguage}      & 10.70 & 29.64 & 41.53 & 9.14 & 28.96 & 44.59 & 27.42 &
                      19.69 & 40.26 & 54.42 & 17.61 & 49.73 & 66.59 & 41.38 \\
PE-RSITR\cite{yuan2023parameter}     & 14.13 & 31.51 & 44.78 & 11.63 & 33.92 & 50.73 & 31.12 &
                      23.67 & 44.07 & 60.36 & 20.10 & 50.63 & 67.97 & 44.47 \\
Full-FT CLIP\cite{radford2021learning}     & 13.54 & 30.83 & 43.46 & 11.55 & 33.14 & 49.83 & 30.39 &
                      24.16 & 47.12 & \textbf{61.28} & \underline{20.40} & 50.53 & 68.54 & 45.33 \\
HarMA\cite{huang2024efficient}        & \underline{16.36} & \underline{34.48} & \underline{47.74} &
                      \underline{12.92} & \textbf{37.17} & \underline{53.07} & \underline{33.62} &
                      \underline{25.81} & \underline{48.37} & 60.61 & 19.92 & \underline{53.27} & \textbf{71.21} & \underline{46.53} \\
\rowcolor{gray!15}
MPS-CLIP (Ours)     & \textbf{18.30} & \textbf{37.42} & \textbf{50.32} &
                      \textbf{13.28} & \underline{37.04} & \textbf{54.73} & \textbf{35.18} &
                      \textbf{27.88} & \textbf{51.11} & \underline{61.06} &
                      \textbf{22.61} & \textbf{56.59} & \underline{71.15} & \textbf{48.40} \\
\bottomrule
\end{tabular}
\vspace{-10pt}
\end{table*}

\subsection{Multi Perspective Learning Objectives}
\label{sec:multi_perspective_loss}

To enhance both global semantic consistency and keyword-driven fine-grained alignment for remote-sensing cross-modal retrieval, we define multi-perspective objectives on the globally aligned pair $(\mathbf{G}_{v},\mathbf{G}_{t})$ and the interaction between the global text embedding and multi-perspective semantic features $(\mathbf{G}_{t},\mathbf{G}_{m})$. This preserves CLIP’s global discrimination while treating local multi-perspective representations as selectable semantic evidence. We select the maximum-response perspective and apply similarity-adaptive weighting ($\lambda_{\text{MPC}}$ and $\lambda_{\text{MPT}}$) to suppress redundant/noisy perspectives, reducing semantic dilution and providing more stable supervision for fine-grained matching. The overall objective is:
\begin{equation}
\mathcal{L}=\mathcal{L}_{\text{Base}}+\lambda_{\text{MPC}}\mathcal{L}_{\text{MPC}}+\lambda_{\text{MPT}}\mathcal{L}_{\text{MPT}}.
\end{equation}

\textbf{Base Loss.}
For the global feature pair $(\mathbf{G}_{v},\mathbf{G}_{t})$, we adopt CLIP bidirectional contrastive loss and a weighted triplet loss to improve cross modal discriminability within a batch and enforce ranking constraints \cite{huang2024efficient}. 

\textbf{Multi Perspective Contrastive Loss.}
In the fine grained branch, we impose a contrastive constraint between the set of multi perspective semantic embeddings $\mathbf{G}_{m}=\{\mathbf{v}_{k}\}_{k=1}^{K}$ produced by MPR and the global text embedding $\mathbf{G}_{t}$. Since directly averaging multiple perspectives mixes complementary information with noise and dilutes local semantics, we introduce the maximum similarity perspective to provide more reliable matching evidence. For any image text pair $(i,j)$, we define the maximum similarity from multi perspective to text as
\begin{equation}
s_{\max}(i,j)=\max_{k}\, s(\mathbf{v}_{k}^{i},\mathbf{G}_{t}^{j}),
\end{equation}
where $s(\cdot,\cdot)$ denotes dot-product similarity. We then replace the image to text similarity term in standard CLIP with $s_{\max}(i,j)$, obtaining the multi perspective contrastive loss
\begin{equation}
\begin{aligned}
\mathcal{L}_{\text{MPC}}
&= -\frac{1}{2B}\sum_{i=1}^{B}
\Bigg[
\log\frac{\exp\big(\tau\, s_{\max}(i,i)\big)}
{\sum_{j=1}^{B}\exp\big(\tau\, s_{\max}(i,j)\big)}
\\
&\qquad\qquad
+
\log\frac{\exp\big(\tau\, s_{\max}(i,i)\big)}
{\sum_{j=1}^{B}\exp\big(\tau\, s_{\max}(j,i)\big)}
\Bigg].
\end{aligned}
\end{equation}
So that local alignment is dominated by the perspective that best matches the text semantics while maintaining the discriminative structure of batch level contrastive learning.

\textbf{Multi Perspective Weighted Triplet Loss.}
To further enhance the separability of multi perspective representations at the ranking level, we reuse the maximum-similarity perspective in Eq.~(16) and directly incorporate it into the weighted triplet loss in Eq.~(12).

The resulting loss is given by
\begin{equation}
\begin{aligned}
\mathcal{L}_{\text{MPT}}
&=
\frac{1}{B}\sum_{i=1}^{B}
w_{t\rightarrow v}^{i}
\big[
m + s_{\max}(i, j)
  - s_{\max}(i, i)
\big]_+
\\
&\quad+
\frac{1}{B}\sum_{i=1}^{B}
w_{v\rightarrow t}^{i}
\big[
m + s_{\max}(j, i)
  - s_{\max}(i, i)
\big]_+.
\end{aligned}
\end{equation}

By integrating the most semantically consistent perspective into the weighted triplet formulation, the model strengthens fine-grained ranking signals while preserving the robustness of the original focal-like weighting scheme.

\section{Experiments}

\begin{figure*}[t]
    \centering
    \includegraphics[width=\linewidth]{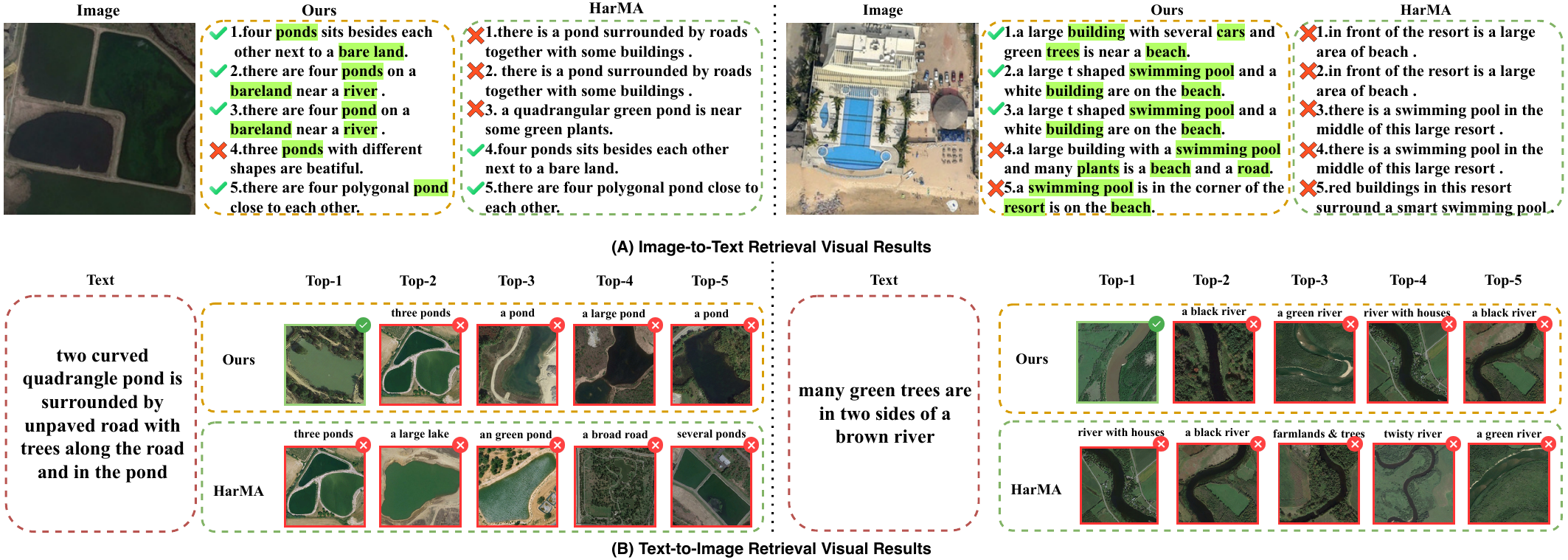}
    \caption{Visual comparison of bidirectional retrieval results between ``Ours" (MPS-CLIP) and HarMA. Green checks ($\checkmark$) and red crosses ($\times$) indicate semantically correct and incorrect retrieval results, respectively.}
    \label{fig:visual_results}
    \vspace{-15pt}
\end{figure*}

\subsection{Datasets and Evaluation Metrics}

We evaluated MPS-CLIP on two standard RSITR datasets: RSICD \cite{lu2017exploring} (10,921 image–text pairs, $224\times224$) and RSITMD \cite{yuan2022exploring} (4,743 pairs, $256\times256$). Each image is paired with five captions. Following HarMA \cite{huang2024efficient}, we split each dataset into train/val/test sets with an 8:1:1 ratio. Performance is reported using Recall@K (R@1, R@5, R@10), and their mean (mR) serves as the overall metric.

\subsection{Implementation Details}

During training, we projected the final features of CLIP into a 512 dimensional embedding space. The temperature parameter in the contrastive loss was fixed to 0.07. We utilized the AdamW optimizer with an initial learning rate of $4 \times 10^{-5}$ and a weight decay of 0.04, employing a linear learning rate decay strategy.  The model was trained for 35 epochs with a batch size of 64 on both datasets. All experiments were conducted on two NVIDIA GeForce RTX 3090 GPUs.

\subsection{Comparison with SOTA Methods}

To comprehensively validate the effectiveness of the proposed MPS-CLIP, we compare MPS-CLIP against various SOTA methods using an identical CLIP backbone to ensure fairness. As shown in Table \ref{tab:rsicd_rsitmd}, MPS-CLIP outperforms competitors on both RSICD and RSITMD. Specifically, on RSICD, it achieves an mR of 35.18, surpassing the strongest competitor HarMA (33.62) by 1.56 points, with R@1 rising to 18.30 (text) and 13.28 (image). On RSITMD, MPS-CLIP boosts mR to 48.40 (+1.87 over HarMA), with notable gains in R@1 for both tasks. These results confirm that our keyword-guided multi-perspective modeling effectively mitigates semantic deficiency and enhances accuracy with minimal overhead.

We provide a qualitative comparison against HarMA. In Image-to-Text tasks (Fig. \ref{fig:visual_results}(A)), MPS-CLIP captures fine-grained details (e.g., ``bare land") more accurately than HarMA's generic descriptions. Similarly, for Text-to-Image retrieval (Fig. \ref{fig:visual_results}(B)), our model successfully identifies ground-truth images at Top-1 for complex queries where HarMA fails. These results visually confirm that our keyword-guided modeling effectively enhances cross-modal semantic alignment.


\subsection{Ablation Studies and Discussion}

In this section, we conducted ablation studies on RSICD dataset to evaluate the contributions of different modules in the MPS-CLIP.

\subsubsection{Ablation on the [CLS] Token in the CLIP}
\label{subsec:ablation_cls}

To verify the effectiveness of the [CLS] token, we constructed a variant where the global representations are derived by calculating the mean of the output text and image features (denoted as $\times$'' in Table \ref{tab:ablation_cls}). This is compared against the standard approach which utilizes the specific [CLS] token embedding (\checkmark). As shown in Table \ref{tab:ablation_cls}, this mean pooling variant results in a clear performance drop. Specifically, R@1 decreases to 14.27 (text) and 10.52 (image). These results demonstrate that simple pooling is less effective at capturing global semantics, whereas the [CLS] token provides a more discriminative representation essential for accurate cross-modal alignment.
\begin{table}[t]
\centering
\setlength{\tabcolsep}{4pt}
\caption{Ablation study on the [CLS] token in CLIP. ``\checkmark'' indicates using the [CLS] token as the global representation for both text and image.}
\label{tab:ablation_cls}
\begin{tabular}{c|ccc|ccc|c}
\toprule
\multirow{2}{*}{\textbf{[CLS]}} & \multicolumn{3}{c|}{\textbf{Text Retrieval}} & \multicolumn{3}{c|}{\textbf{Image Retrieval}} & \multirow{2}{*}{\textbf{mR}} \\
\cline{2-7}
& \textbf{R@1} & \textbf{R@5} & \textbf{R@10} & \textbf{R@1} & \textbf{R@5} & \textbf{R@10} &  \\
\midrule
$\times$      & 14.27 & 32.75 & 44.65 & 10.52 & 34.75 & 51.25 & 31.37 \\
\checkmark    & \textbf{18.30} & \textbf{37.42} & \textbf{50.32} & \textbf{13.28} & \textbf{37.04} & \textbf{54.73} & \textbf{35.18} \\
\bottomrule
\end{tabular}
\vspace{-10pt}
\end{table}

\subsubsection{Ablation on G$^2$A Adapter Components}

Table \ref{tab:ablation_attn_gate} evaluated the contributions of Attn and Gate in the G$^2$A adapter. Adding Attn improves mR from 34.34 to 34.59, confirming that global interaction in compressed space aids adaptation. Introducing Gate yields a larger gain (to 34.99), highlighting the importance of adaptive residual blending for preserving pretrained knowledge. Combining both achieves the best performance, demonstrating their complementarity: Attn enhances global dependency modeling while Gate stabilizes optimization, all with minimal parameter overhead.

\begin{table}[t]
\centering
\footnotesize
\setlength{\tabcolsep}{4pt}
\caption{Comparison with different combinations of the \textit{Attn} and \textit{Gate} components in the G$^2$A adapter (the bottleneck projections are always enabled). Parameters and FLOPs for a single-layer adapter are also reported. The symbol ``\checkmark'' indicates that the corresponding module is enabled.}
\label{tab:ablation_attn_gate}

\resizebox{\linewidth}{!}{%
\begin{tabular}{cc|ccc|ccc|c|cc}
\toprule
\multicolumn{2}{c|}{\textbf{Modules}} &
\multicolumn{3}{c|}{\textbf{Text Retrieval}} &
\multicolumn{3}{c|}{\textbf{Image Retrieval}} &
\multirow{2}{*}{\textbf{mR}} &
\multirow{2}{*}{\textbf{Params}} &
\multirow{2}{*}{\textbf{FLOPs}} \\
\cline{1-8}
\textbf{Attn} & \textbf{Gate} &
\textbf{R@1} & \textbf{R@5} & \textbf{R@10} &
\textbf{R@1} & \textbf{R@5} & \textbf{R@10} &  &  &  \\
\midrule
$\times$      & $\times$      & 17.02 & 34.95 & 48.95 & 13.16 & 37.57 & 54.38 & 34.34 & \textbf{0.49} & 0.05 \\
\checkmark    & $\times$      & 17.29 & 37.15 & 48.67 & 13.23 & 37.11 & 54.09 & 34.59 & 0.51          & 0.05 \\
$\times$      & \checkmark    & 17.47 & 36.23 & 49.31 & \textbf{13.98} & \textbf{37.99} & \textbf{54.93} & 34.99 & \textbf{0.49} & 0.05 \\
\checkmark    & \checkmark    & \textbf{18.30} & \textbf{37.42} & \textbf{50.32} & 13.28 & 37.04 & 54.73 & \textbf{35.18} & 0.51 & 0.05 \\
\bottomrule
\end{tabular}%
}
\end{table}

\subsubsection{Ablation Study on MPR module}

\begin{table}[t]
\centering
\setlength{\tabcolsep}{4pt}
\caption{Ablation study of the Multi-Perspective Representation (MPR) module. ``\checkmark'' indicates that MPR module is enabled.}
\label{tab:ablation_mpr}
\begin{tabular}{c|ccc|ccc|c}
\toprule
\multirow{2}{*}{\textbf{MPR}} & \multicolumn{3}{c|}{\textbf{Text Retrieval}} & \multicolumn{3}{c|}{\textbf{Image Retrieval}} & \multirow{2}{*}{\textbf{mR}} \\
\cline{2-7}
& \textbf{R@1} & \textbf{R@5} & \textbf{R@10} & \textbf{R@1} & \textbf{R@5} & \textbf{R@10} &  \\
\midrule
$\times$  & 17.38 & 35.41 & 48.95 & \textbf{13.36} & \textbf{37.15} & \textbf{54.78} & 34.50 \\
\checkmark & \textbf{18.30} & \textbf{37.42} & \textbf{50.32} & 13.28 & 37.04 & 54.73 & \textbf{35.18} \\
\bottomrule
\end{tabular}
\vspace{-15pt}
\end{table}

As shown in Table \ref{tab:ablation_mpr}, incorporating MPR boosts overall performance, raising mR from 34.50 to 35.18. Notably, text retrieval sees significant gains (R@1: 17.38 → 18.30), indicating that MPR effectively aggregates keyword-guided sub-perspectives to reduce noise and enhance fine-grained alignment. While image retrieval shows slight fluctuations, we attribute this to MPR's primary focus on refining text-driven local semantics, which benefits text-to-image matching more directly.

\subsubsection{Analysis of Multi-Perspective Loss Functions}
Table \ref{tab:ablation_loss} shows that the baseline objective ($\mathcal{L}_{\text{Base}}$) achieves an mR of 34.50. Incorporating $\mathcal{L}_{\text{MPC}}$ improves mR to 34.65 by selecting maximum-response perspectives to reduce noise, while $\mathcal{L}_{\text{MPT}}$ reaches 34.74 by emphasizing text-consistent perspectives via adaptive reweighting. Combining both yields the best performance (mR 35.18), demonstrating that hard perspective selection and adaptive ranking supervision are complementary strategies for robust cross-modal alignment.

\begin{table}[t]
\centering
\setlength{\tabcolsep}{4pt}
\caption{Ablation study results of different loss combinations.}
\label{tab:ablation_loss}

\resizebox{\linewidth}{!}{%
\begin{tabular}{ccc|ccc|ccc|c}
\toprule
\multicolumn{3}{c|}{\textbf{Losses}} &
\multicolumn{3}{c|}{\textbf{Text Retrieval}} &
\multicolumn{3}{c|}{\textbf{Image Retrieval}} &
\multirow{2}{*}{\textbf{mR}} \\
\cline{1-9}
\textbf{$\bm{\mathcal{L}_{\text{Base}}}$} &
\textbf{$\bm{\mathcal{L}_{\text{MPC}}}$} &
\textbf{$\bm{\mathcal{L}_{\text{MPT}}}$} &
\textbf{R@1} & \textbf{R@5} & \textbf{R@10} &
\textbf{R@1} & \textbf{R@5} & \textbf{R@10} & \\
\midrule
\checkmark &            &            &
17.38 & 35.41 & 48.95 &
13.36 & \textbf{37.15} & \textbf{54.78} & 34.50 \\
\checkmark & \checkmark &            &
17.38 & 36.23 & 49.77 &
\textbf{13.92} & 37.05 & 53.56 & 34.65 \\
\checkmark &            & \checkmark &
16.28 & 35.59 & 48.40 &
12.48 & 36.72 & 53.40 & 34.74 \\
\checkmark & \checkmark & \checkmark &
\textbf{18.30} & \textbf{37.42} & \textbf{50.32} &
13.28 & 37.04 & 54.73 & \textbf{35.18} \\
\bottomrule
\end{tabular}%
}
\vspace{-8pt}
\end{table}

\section*{Conclusions}

In this paper, we presented MPS-CLIP, a novel parameter-efficient framework designed to bridge the granularity gap in RSITR. To overcome the limitations of traditional coarse-grained global alignment, MPS-CLIP shifts the paradigm towards a semantic-driven, multi-perspective alignment strategy. By seamlessly integrating LLMs for keyword mining and SamGeo for region generation, our approach explicitly captures fine-grained local semantics and complex spatial relationships inherent in remote sensing imagery. Furthermore, the proposed G$^2$A adapter and MPR module ensure robust feature fusion and efficient adaptation of the CLIP backbone. Extensive experiments on the RSICD and RSITMD benchmarks demonstrate that MPS-CLIP achieves new SOTA performance, significantly outperforming existing competitive methods and full fine-tuning baselines. These results validate the efficacy of multi-perspective modeling in mitigating semantic ambiguity. In future work, we plan to extend this keyword-guided interaction paradigm to more dense prediction tasks, such as remote sensing visual grounding and captioning.

\bibliographystyle{IEEEbib}
\bibliography{icme2026references}

\end{document}